\documentclass[runningheads]{llncs}

\usepackage{eccv}

\usepackage{eccvabbrv}

\usepackage{graphicx}
\usepackage{booktabs}
\usepackage{multirow}

\usepackage[accsupp]{axessibility}  % Improves PDF readability for those with disabilities.

\usepackage{hyperref}

\usepackage{orcidlink}

\begin{document}

% ---------------------------------------------------------------
    \title{PoTATO: A Dataset for Analyzing Polarimetric Traces of Afloat Trash Objects} 

\titlerunning{PoTATO: Polarimetric Image Dataset for Detecting Floating Objects}

\author{Luis F. W. Batista\inst{1,2}\orcidlink{0009-0006-2464-5841} \and
Salim Khazem\inst{2,3}\orcidlink{0000-0001-5958-6120} \and
Mehran Adibi\inst{2}\orcidlink{0009-0008-5817-5731}\and
\\ Seth Hutchinson\inst{1}\orcidlink{0000-0002-3949-6061} \and Cedric Pradalier\inst{2}\orcidlink{0000-0002-1746-2733}}

\authorrunning{L.F.W.~Batista et al.}

\institute{Georgia Institute of Technology, Atlanta, USA \\
\and GeorgiaTech Europe - IRL2958 GT-CNRS, Metz, FR \and CentraleSupelec, Metz, FR}

\maketitle
\begin{abstract}
Plastic waste in aquatic environments poses severe risks to marine life and human health. Autonomous robots can be utilized to collect floating waste, but they require accurate object identification capability. While deep learning has been widely used as a powerful tool for this task, its performance is significantly limited by outdoor light conditions and water surface reflection. Light polarization, abundant in such environments yet invisible to the human eye, can be captured by modern sensors to significantly improve litter detection accuracy on water surfaces. With this goal in mind, we introduce PoTATO, a dataset containing 12,380 labeled plastic bottles and rich polarimetric information. We demonstrate under which conditions polarization can enhance object detection and, by providing raw image data, we offer an opportunity for the research community to explore novel approaches and push the boundaries of state-of-the-art object detection algorithms even further. Code and data are publicly available at  \url{https://github.com/luisfelipewb/PoTATO/tree/eccv2024}.

\end{abstract}

\section{Introduction}
\label{sec:intro}

\begin{figure}[t]
    \centering
    \includegraphics[width=\linewidth]{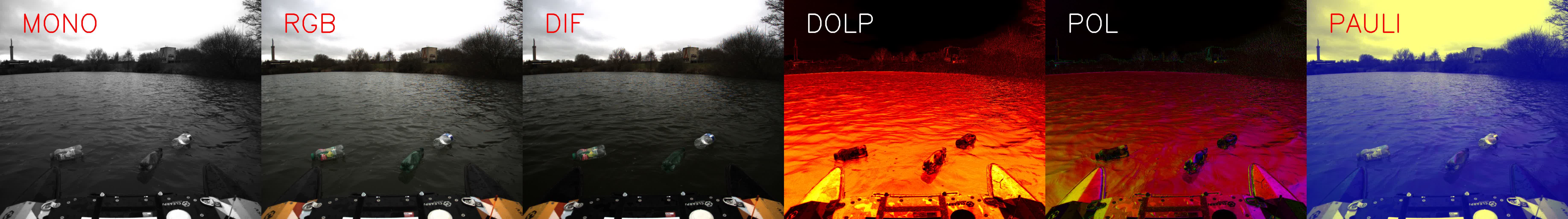}
    \caption{Different modalities that can be extracted from the dataset: Grayscale (MONO), Color (RGB), Color image with diffuse-only reflections (DIF), Degree of Linear Polarization (DOLP), pseudo-color image combining degree and angle of polarization (POL), and pseudo-color Pauli-inspired image (PAULI)}
    \label{fig:modalities}
\end{figure}

% Floating plastic problem
Plastic pollution is a global threat known for its capability of damaging aquatic life, ecosystems, and even human health \cite{van2020plastic}. Plastic is inherently designed to last a long time, but the weathering processes cause fragmentation into smaller particles that can reach remote locations on the planet \cite{macleod2021global}. The plastic accumulation in such areas is poorly reversible because the cleanup actions are infeasible and the natural removal process is slow. According to Chamas et al., \cite{chamas2020degradation}, the only known solution to decrease plastic accumulation is to reduce its emissions.

To prioritize actions that can prevent plastic from reaching the natural environment, it is necessary to identify the transport mechanisms and how it is reaching the ocean. van Emmerik and Schwarz \cite{van2020plastic} have identified a lack of thorough understanding of how plastic is transported from land to aquatic systems, especially going through rivers. Most observation-based studies rely on existing infrastructure such as bridges and have limited coverage for a detailed understanding of how trash is reaching the ocean. Such systems can aid monitoring the trash flow, but are rarely capable of catching it.

% Object detection introduction and bridge to multiple modalities
Machine learning, specifically through object detection, has been widely utilized to automate the quantification of plastic in bodies of water \cite{politikos2023using}. The field of object detection, considered one of the most fundamental tasks in computer vision, has made remarkable progress over the past decade, with neural networks emerging as a powerful tool for identifying and classifying objects in various contexts \cite{zou2023object}. Despite the advancements in object detection techniques, challenges remain, especially in outdoor environments where lighting conditions can vary significantly creating bright sunlight, shadows, glare, and reflections on bodies of water. To address these challenges, the fusion of sensors of different modalities has been explored in different studies \cite{blin2021polarlitis, cheng2021flow}.

% How polarization can be useful. 
Modern camera sensors can have the Bayer array integrated with microgrid polarizers, allowing to capture polarized and color images simultaneously \cite{wang2022principle}. This technology has led to increased interest in the computer vision and robotics research community to leverage polarimetric information for enhanced perception systems. According to Andreou et al. \cite{andreou2002polarization}, utilizing polarization sensitivity can aid object detection by increasing the visual contrast of concealed objects. This is particularly relevant in outdoor environments and bodies of water, where two natural sources of polarization are abundant: scattered light and surface reflections \cite{foster2018polarisation}. Integrating polarimetric data into object detection holds immense potential to improve system performance and reliability.

% Paper contribution
The main contributions of our work are:

\textbf{1.} The first labeled dataset with raw, pixel-wise-aligned polarimetric and chromatic data for object detection.

\textbf{2.} An analysis of polarized light's key physical properties in outdoor, water-rich environments, highlighting its potential for object detection.

\textbf{3.} A baseline comparison of three well-known object detection algorithms across six image modalities, showcasing scenarios where polarized images surpass color images in detection efficacy.

The PoTATO dataset offers value beyond the object detection problem due to its versatility. Providing the raw data allows for research across various domains, including microgrid polarized image demosaicking and multi-modality fusion.
\section{Related Work}
\label{sec:related_work}

%Overview
\textbf{Deep Learning-based object detection on water surface}.
Over the past three years, there has been a significant increase in the number of publications utilizing AI techniques on marine macrolitter datasets \cite{politikos2023using}. Despite the recent advancements, a lack of Deep Learning models with satisfactory generalization capability to detect trash on bodies of water has been identified, and further research is necessary to develop robust models that can overcome the challenges posed by the diverse geographical and environmental conditions \cite{jia2023deep}. Additionally, the vast majority of existing approaches rely on satellite images, drone images, or cameras fixed on bridges. While these are valuable for monitoring and quantifying litter, they can not be easily integrated into robotic systems that are capable of autonomously collecting the identified waste.

% Challenges
The detection of floating objects is often hindered by challenges such as the varying outdoor illumination conditions and sun glares and the reflections on the water's surface, as highlighted in \cite{zhang2021survey}. To address these challenges, recent research has focused on the fusion of sensors employing different modalities. For example, in \cite{iqbal2022object}, a combination of a long-wave infrared polarimeter camera and a visible wavelength optical camera was utilized, while in \cite{cheng2021flow}, a millimeter wave radar was integrated with a camera. However, synchronizing and aligning measurements from different sensors introduces additional complexity and can potentially impact the performance of the model.

\textbf{Object Detection using polarized images}.
The development of microgrid polarization sensors has leveraged polarized images for various applications, including, 3D reconstruction \cite{dave2022pandora}, pose prediction 
 \cite{gao2022polarimetric}, and reflection separation \cite{lei2020polarized}. More specifically for object detection, a deep learning approach utilizing polarization images was proposed for car detection and the results showed promise in addressing illumination and reflection issues \cite{ratliff2011detection}. In a subsequent study, Blin et al. compared models trained independently on a range of polarimetric encoded inputs and RGB images, demonstrating improved results compared to RGB-based detection \cite{blin2019adapted}. Next, various multimodal fusion schemes and combinations of chromatic and polarimetric features were investigated  \cite{blin2021multimodal}. While the results are encouraging, the utilization of two independent sensors and the lack of pixel-wise correspondence between the images, make early-fusion approaches more challenging.

Recently, polarization information has been successfully combined with grayscale intensity to improve the results of semantic segmentation of transparent objects \cite{kalra2020deep}. In their work, an attention-based early-fusion approach is proposed, but it does not use color information and the indoor application lacks the complexity of the outdoor illumination conditions. 

Current research shows the strong potential of using polarized images for object detection in adverse illumination conditions but lacks a public dataset including raw data and pixel-wise alignment.

\textbf{Related datasets}.
Environmental protection has been receiving increased attention, resulting in the proposal of several datasets targeting litter detection. The analysis performed by Politikos et al. \cite{politikos2023using} provides a database for multiple datasets. Nonetheless, very few images are captured from the point of view of the vessel, making them hard to use in robotic systems capable of collecting waste autonomously. Vessel-mounted cameras have already been employed to quantify the presence of plastic in the ocean, \cite{de2021quantifying} and a large dataset comprising colored images was utilized. Nonetheless, public access to it was not identified.

The Flow-Img dataset \cite{cheng2021flow} presents 2000 real-world labeled images captured from an Unmanned Surface Vessel (USV) and shares a similar application. Their study demonstrates that existing alternatives, such as TrashNet \cite{wang2018bottle}, and TACO \cite{proencca2020taco} do not effectively generalize to complex environments and diverse perspectives and emphasize the need for expanding datasets containing images captured from the point of view of the vessel.

The PolarLITIS dataset  \cite{blin2021polarlitis} was used specifically to compare the performance of color and polarization images. However, it is intended for road scenes without abundant water surface reflection, and the data was acquired using different sensors, leading to pixel-wise misalignment.

To the best of our knowledge, PoTATO is the first dataset containing raw polarized and colored images with pixel-wise alignment for object detection.

\section{PoTATO Dataset}

\begin{figure}[bth]
    \centering
    \includegraphics[width=0.7\linewidth]{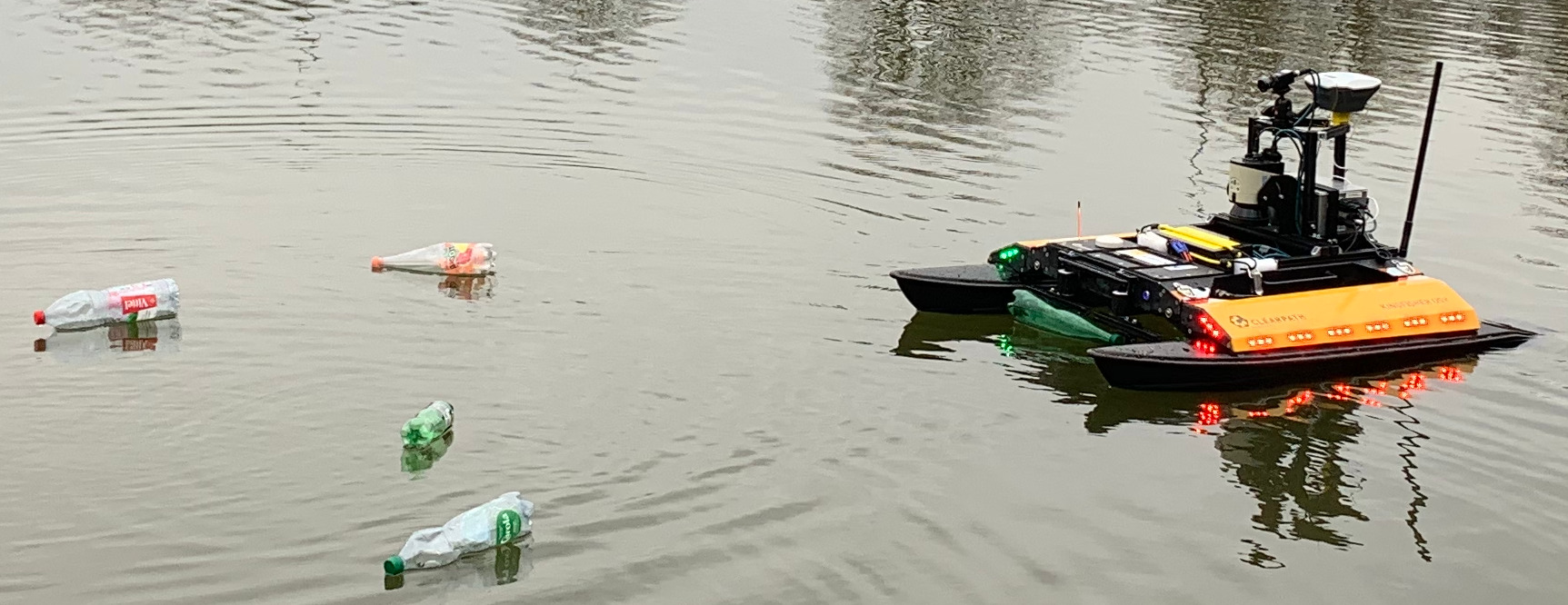}
    \caption{Polarized camera mounted on the Kingfisher USV}
    \label{fig:acquisition-platform}
\end{figure}

% \subsection{Acquisition Platform}
To acquire the images, the camera Triton $5.0 MP$  TRI050S1-QC was used. This camera utilizes the SONY sensor IMX264MYR, capable of capturing colored images and polarization with resolution of $2448\times2048$ pixels. It was paired with a lens with a focal length of $6 mm$, providing an angle of view of $80.8^{\circ}(H) \times 61.6^{\circ}(V)$. The camera was mounted on a Kingfisher USV from Clearpath Robotics. It was positioned facing forward with a downward angle to achieve the best field of view for observing the floating objects in the water up to the vessel's approach, as shown on \cref{fig:acquisition-platform}.

% \subsection{Data Collection and Processing}
Data collection took place on Lake Symphonie across seven distinct days, each characterized by varying weather as detailed in \cref{table:dataset_numbers}. Each recording session comprised several short clips to capture a broad spectrum of conditions and assure the variability of the polarimetric information. By approaching bottles from various directions and relocating them across the lake, we ensured a diversity of backgrounds, lighting, and relative sun positions. The boat was operated manually, and the images were recorded at two frames per second. Due to the variable light conditions in the outside environment, the auto-brightness function of the camera was enabled. All images containing bottles also include varied background features. Since background-only images minimally enhance detection, the dataset exclusively comprises images with at least one bottle. To allow flexibility and enable the exploration of different approaches, we recorded our dataset in the raw image format with a resolution of $2448\times2048$ pixels, and six different modalities depicted in \cref{fig:modalities} were extracted using the pipeline explained in \Cref{subsec:pipeline}.

\begin{table}
    \captionof{table}{Dataset Statistics and Weather Conditions}
    \centering
    \begin{tabular}{c|r|r|c}
    \textbf{Day} & \textbf{Images}  & \textbf{Labels}  & \textbf{Weather} \\
    \hline
    01 & 27      & 81               & Sunny             \\
    02 & 114     & 414              & Sunny             \\
    03 & 462     & 1450             & Sunny             \\
    04 & 1658    & 4392             & Sunny             \\
    05 & 902     & 2096             & Partially Cloudy  \\
    06 & 459     & 787              & Cloudy            \\
    07 & 978     & 3160             & Cloudy            \\
    \hline
          & 4600    & 12380            &                  
    \end{tabular}
    \label{table:dataset_numbers}
\end{table}

% \subsection{Annotation}

The dataset has a single class \textit{bottle} and annotation was performed using the Label Studio tool \cite{Label_Studio} using machine-learning-aided pre-annotations. The bounding boxes were drawn tightly, encapsulating only the bottle and not including its reflection on the water surface. During labeling, the RGB images were used due to its easy interpretability to the annotator. Nonetheless, the raw image format, potentially contained more features that could allow identifying objects that were not visible on the colored image only. To avoid introducing this bias in the dataset, the annotation process was carried out carefully only by individuals who participated in the image recording. This allowed using the previous knowledge of the locations of the bottles to minimize the chance of False Negatives and False Positives in the ground truth. In addition to that, the images were labeled sequentially, enabling the annotator to leverage temporal information to create precise annotations.

% \subsection{Dataset Split and Statistics}

\begin{figure}[bht]
\begin{center}
\includegraphics[width=0.8\linewidth]{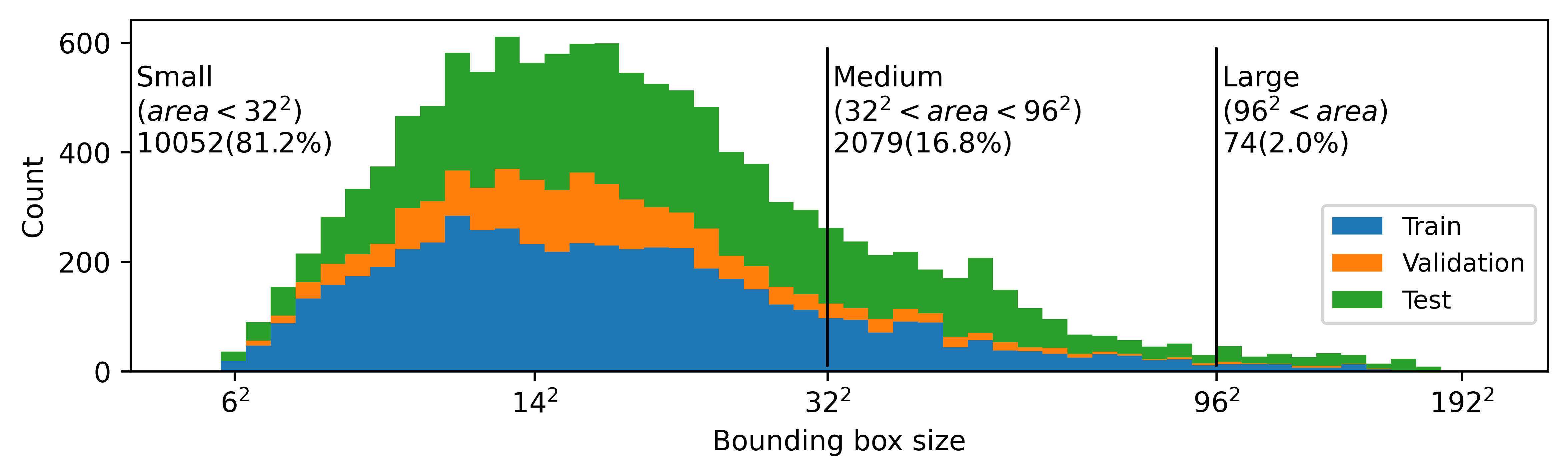}
\end{center}
   \caption{Size distribution of bounding boxes.}
\label{fig:small-bboxes}
\end{figure}

The dataset was split into train, validation, and test sets with 2000 $(43.5\%)$, 600 $(13.0\%)$ and, 2000 $(43.5\%)$ images respectively. In our experiments, we kept a high number of images in the test set to increase the statistical significance of our evaluations. This was important due to the high variety of illumination conditions present on the dataset. During the split of the dataset we considered that consecutive images can have higher similarity and to prevent any leak from training to test set, the images from each recording session were split sequentially into train, validation and test sets. Each image contains between $1$ and $8$ labels. In total, the dataset has $12,380$ annotated plastic bottles and the bounding box size distribution is shown \cref{fig:small-bboxes}.

\section{Overview of Light Polarization Theory}
\label{sec:theory}

This section provides a theoretical background on light polarization and its application to the PoTATO dataset. We first discuss natural sources of polarized light in aquatic environments and their impact on image formation. Next, we explore the mathematical framework for extracting polarimetric data from raw sensor measurements. Finally, we present an overview of the pipeline used to obtain the desired polarimetric visualizations.

\subsection{Sources of Polarization}
\label{subsec:polarization-sources}

Water bodies in outdoor environment have abundance of two natural sources of polarized light: reflection and skylight scattering \cite{foster2018polarisation}. The first occurs when the light reflected on a dielectric material becomes linearly horizontally polarized, and its degree of polarization depends on the angle of incidence relative to the surface normal. The angle where the maximum polarization occurs is known as Brewster's angle ($\theta_{i_{B}}$) and, from Snell's Law, it can be related to the refractive index of the two mediums with the simple expression shown in \cref{eq:brewster_angle}. \cite{goldstein2017polarized}

\begin{equation}
    \theta_{i_{B}} = \arctan(\frac{\eta_{water}}{\eta_{air}})
\label{eq:brewster_angle} 
\end{equation}

\begin{figure}[htbp]
    \centering
    \begin{subfigure}{0.35\linewidth}
    \includegraphics[width=\linewidth]{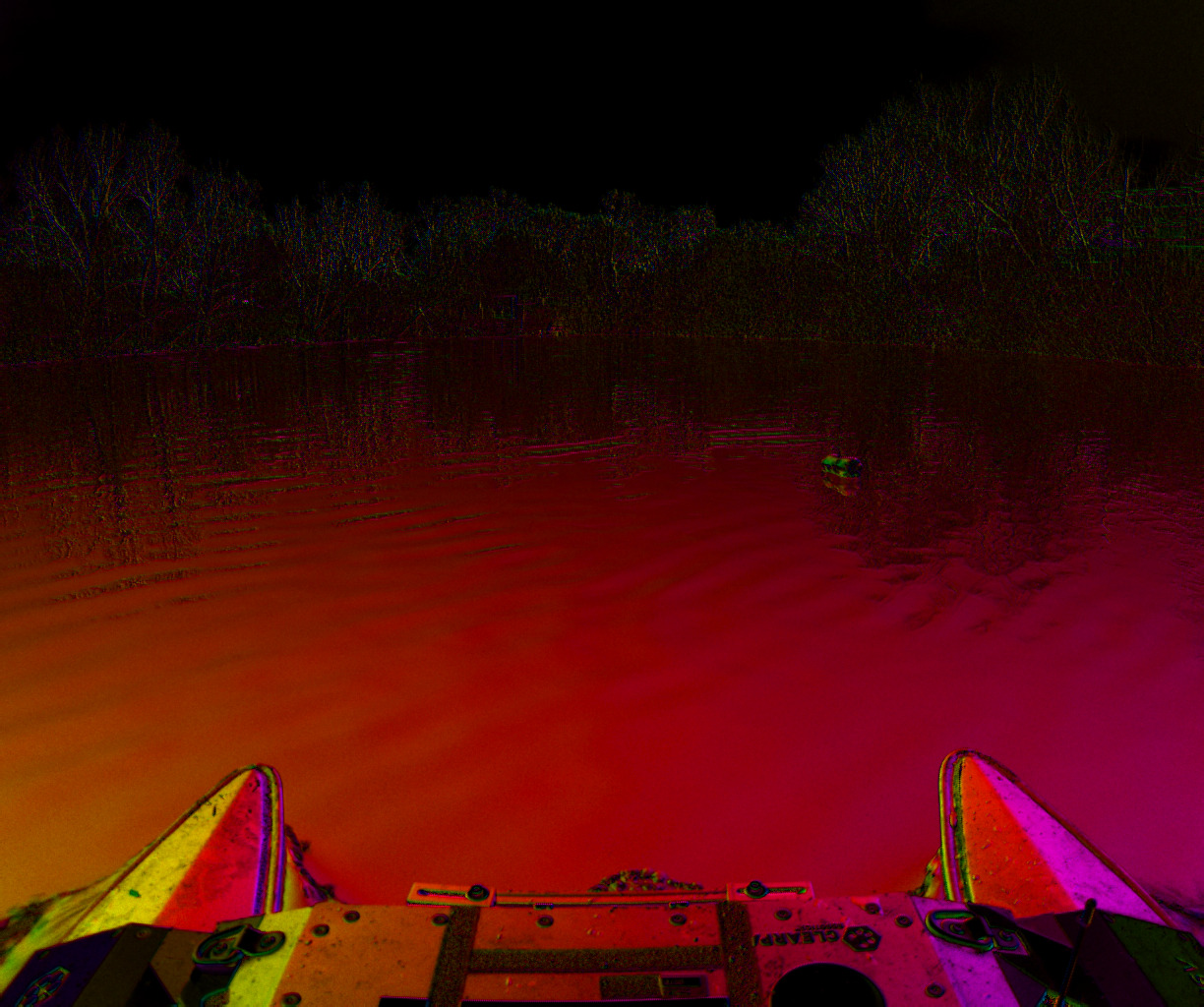}
    \caption{Polarization by reflection on water surface during a cloudy day}
    \label{fig:cloudy-sample}
    \end{subfigure}
    \begin{subfigure}{0.35\linewidth}
    \includegraphics[width=\linewidth]{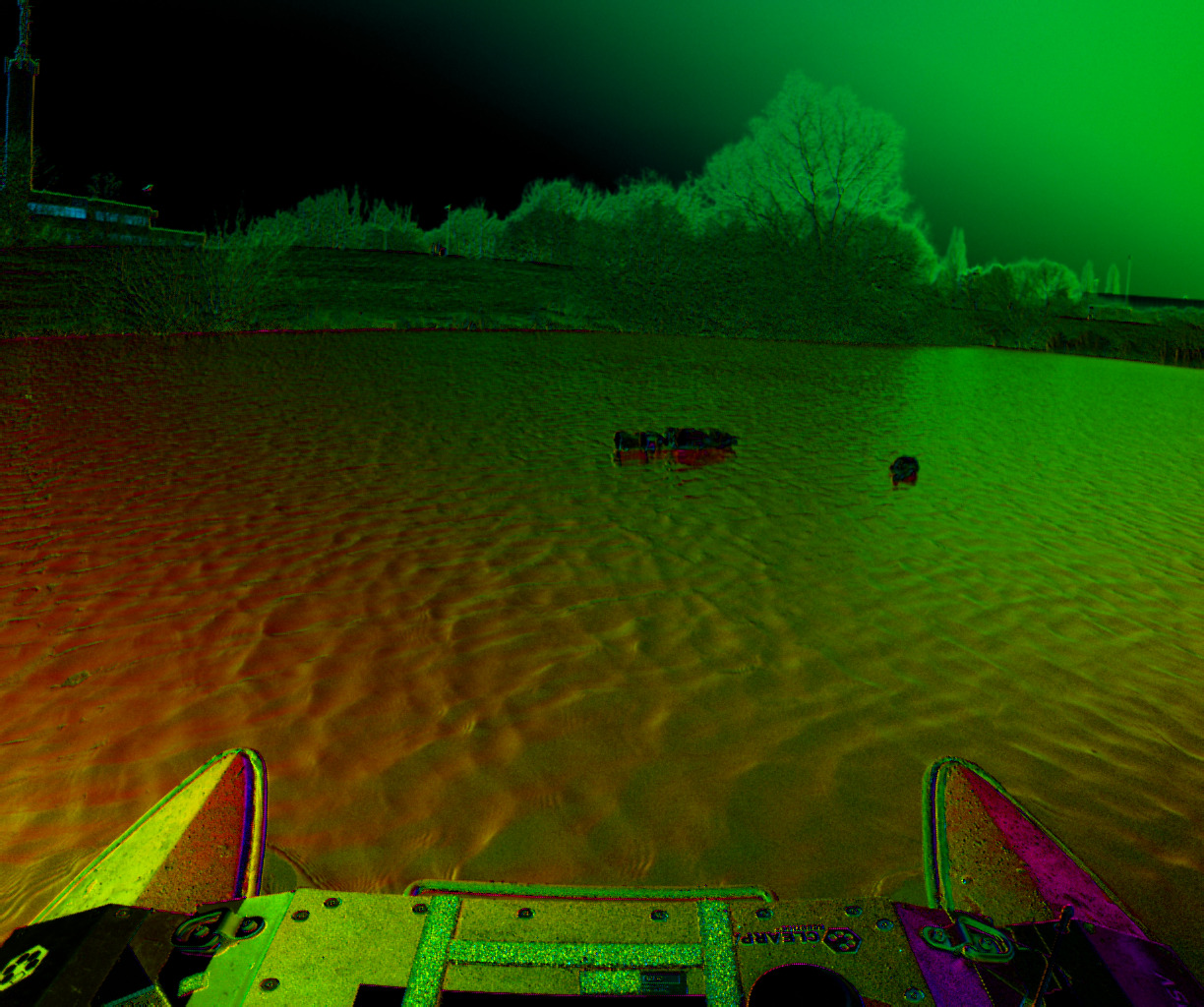}
    \caption{Rayleigh scattering of skylight during a day with clear sky}
    \label{fig:sunny-sample}
    \end{subfigure}
    \caption{Main natural sources of light polarization}
    \label{fig:polarization-sources}
\end{figure}

Given the refractive index of water and air $\eta_{air}  = 1$ and $\eta_{water} \approx 1.33$, The Brewster's angle  $\theta_{i_{B}}$ can be calculated and has a value of approximately $53^{\circ}$. The camera is installed in the vessel at a height of $75 cm$, therefore, the region with the strongest polarization by reflection is located around $1 m$ distance. Polarization by reflection is predominant in cloudy weather and shown in \cref{fig:cloudy-sample}. 

The second occurs when light enters the atmosphere and scatters toward an observer through interactions with atoms and molecules. The degree of polarization increases with angular deviation, reaching its peak at $90^{\circ}$. This phenomenon is known as Rayleigh scattering and produces diverse polarization patterns depending on the sun's position relative to the observer \cite{foster2018polarisation}. Skylight polarization is abundant under clear skies and is shown in \cref{fig:sunny-sample}.

\subsection{Extracting Polarimetric Data}
\label{subsec:pipeline}

\begin{figure}[htb]
\centering
% Left Column
\begin{minipage}{0.35\linewidth}
    % \fbox{
        \begin{subfigure}{\linewidth}
        \includegraphics[width=\linewidth]{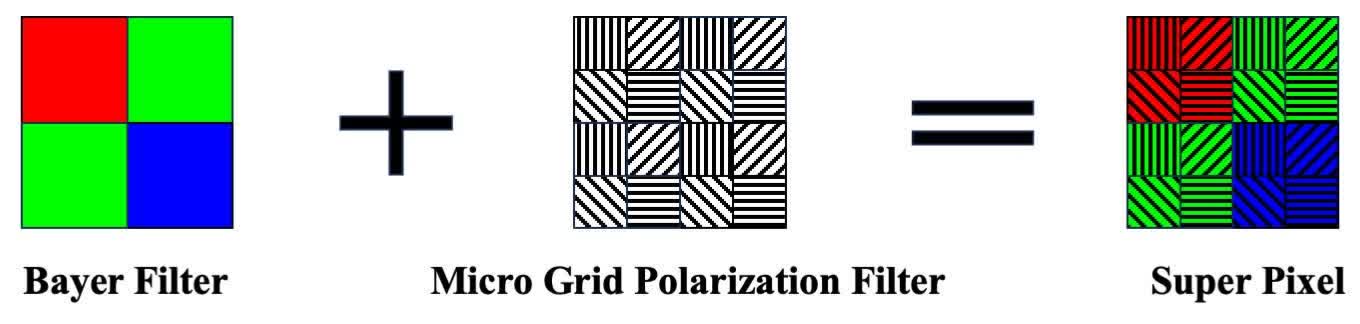}
        \caption{Super Pixel}
        \label{fig:sensor}
        \end{subfigure}
    % }
    \hfill
    % \fbox{
    \begin{subfigure}{\linewidth}
        \includegraphics[width=\linewidth]{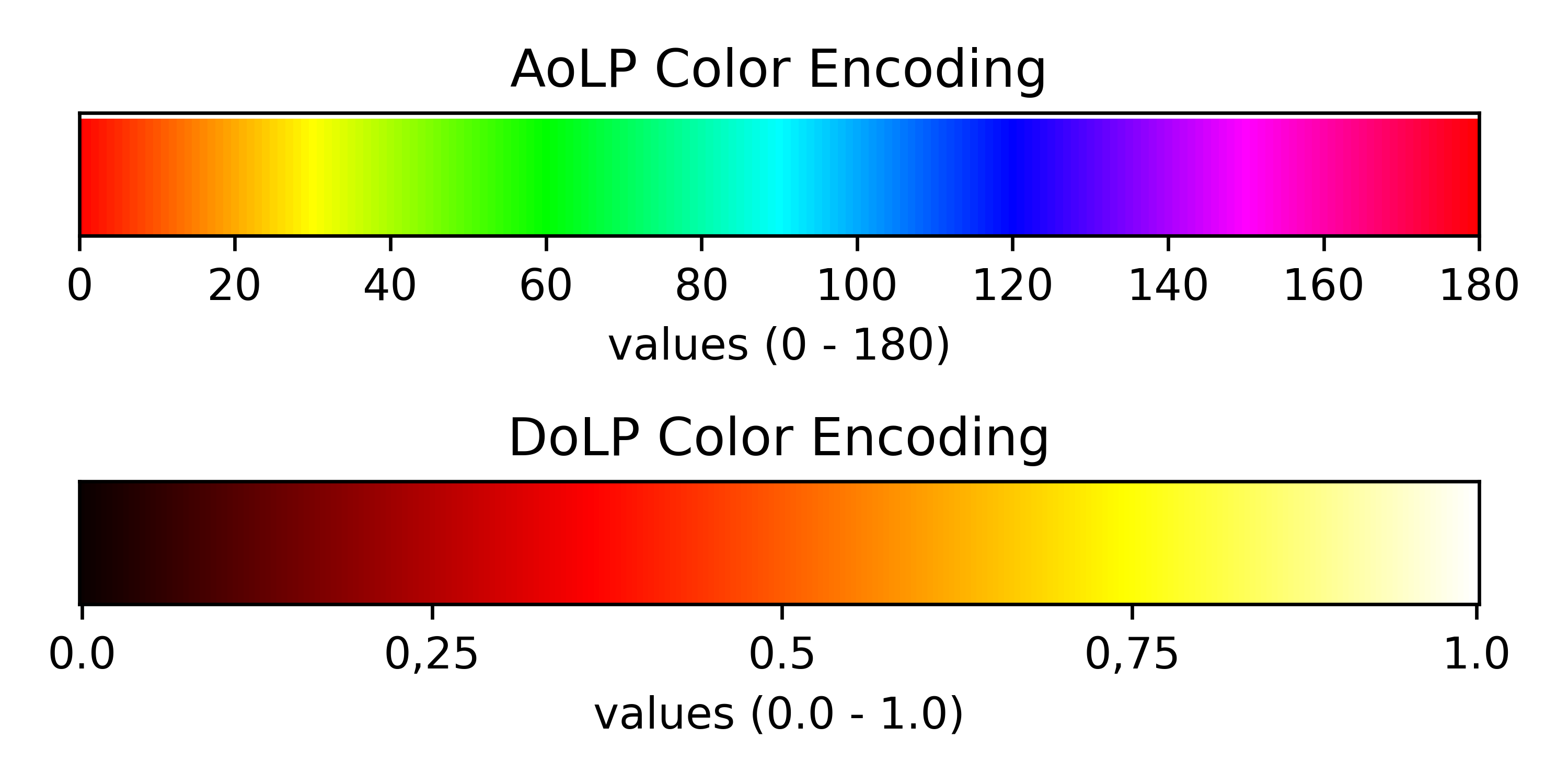}
        \caption{Pseudo color encoding}
        \label{fig:color-encoding}
    \end{subfigure}
    % }
\end{minipage}
\hfill
% Right Column
\begin{minipage}{0.61\linewidth}
    % \fbox{
    \begin{subfigure}{\linewidth}
        \includegraphics[width=\linewidth]{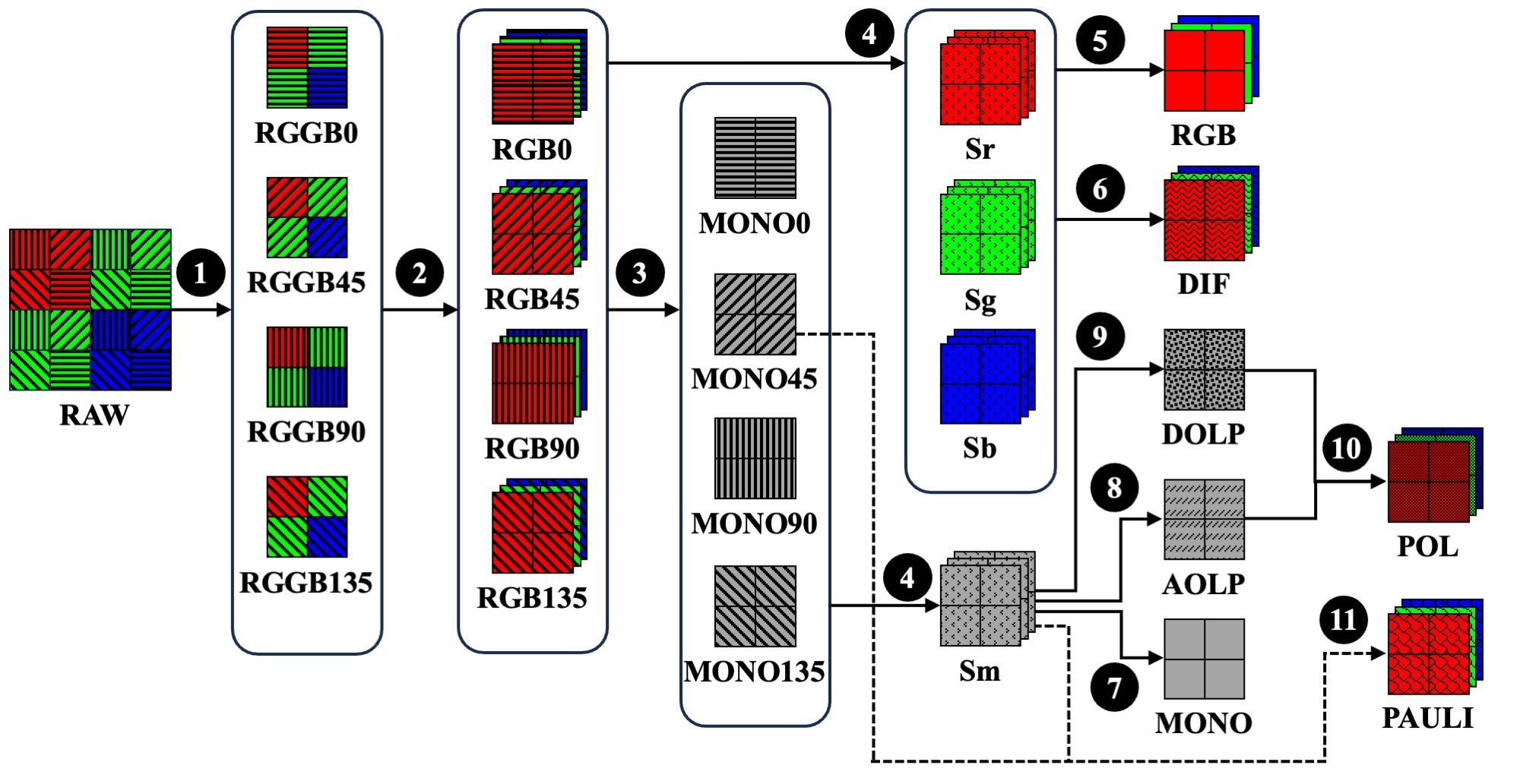}
        \caption{Extraction Pipeline}
        \label{fig:pipeline}
    \end{subfigure}
    % }
\end{minipage}

\caption{Super Pixel, Extraction Pipeline and Color Encoding}
\label{fig:microgrid-polarization-sensor}
\end{figure}

The PoTATO dataset provides the raw data obtained from the sensor in a single-channel image. The sensor features a $2 \times 2$ microgrid polarizer capable of measuring the intensity of light after it has passed through a linear polarizer oriented at four distinct angles $(0^{\circ}, 45^{\circ}, 90^{\circ}, 135^{\circ})$. 
These intensity measurements are denoted as $I_{0}$, $I_{45}$, $I_{90}$, and $I_{135}$ respectively and provide the necessary data to calculate the Stokes parameters. By superimposing the polarization filter with the Bayer array, a super pixel structure is created allowing the measurement of polarization for multiple colors (\cref{fig:sensor}). This integration with varying spatial positions of pixels introduces the possibility of edge artifacts \cite{ratliff2009interpolation} but ensures completely synchronized measurements, with high stability.

The Stokes parameters, which are calculated using the intensity measurements mentioned above, provide a complete description of the polarization state of the electromagnetic radiation \cite{goldstein2017polarized}. These parameters are derived from specific combinations of the measured intensities (\cref{eq:stokes}) and are critical for understanding the polarization properties of the light in the image. As the circular polarization is rare in nature and the parameters $I_L$ and $I_R$ cannot be measured by this sensor, the Stokes parameter $S_3$ is considered to be zero and omitted in subsequent equations.

\begin{equation}
\label{eq:stokes} 
\begin{bmatrix} 
% S_0 & S_1 & S_2 & S_3
S_0 \\ S_1 \\ S_2 \\ S_3
\end{bmatrix}%^{T}
=
\begin{bmatrix}
% I_{0}+I_{90};& I_{0} - I_{90};& I_{45} - I_{135};& I_R - I_L
I_{0}+I_{90}\\ I_{0} - I_{90}\\ I_{45} - I_{135} \\ I_R - I_L
% \frac{I_{0}+I_{45}+I_{90}+I_{135}}{2}\\ I_{0} - I_{90}\\ I_{45} - I_{135} \\ $I_R$ - $I_L$
\end{bmatrix}%^{T}
\end{equation}
Using the stokes parameters, it is possible to derive additional parameters. In our study, we utilize the Degree of Linear Polarization ($DoLP$), Angle of Linear Polarization ($AoLP$), and the Intensity of Diffuse Reflection ($I_{dif}$), with their respective equations extracted from \cite{goldstein2017polarized}. The $DoLP$ indicates the proportion of linearly polarized light with values ranging from $0$ (completely unpolarized) to $1$ (completely polarized) \cref{eq:dolp}.
The $AoLP$ represents the orientation of the electric field vector with values ranging from $0^{\circ}$ to $180^{\circ}$ (\cref{eq:aop}). Finally, $I_{dif}$ can be estimated through \cref{eq:dif}. It filters out specular reflection by considering that polarization is created by reflection. This  technique allows for reducing the intensity of reflections on the water surface.

\begin{equation}
    DoLP = \frac{\sqrt{S_1^{2} +S_2^{2}}}{S_0}
\label{eq:dolp} 
\end{equation}
\begin{equation}
    AoLP = \frac{1}{2} arctan(\frac{S_2}{S_1})
\label{eq:aop} 
\end{equation}
\begin{equation}
    I_{dif} = \frac{S_0 - \sqrt{S_1^{2}+S_2^{2}}}{2}
\label{eq:dif}
\end{equation}

\subsection{Implemented Pipeline}

In our work, we implemented the pipeline shown in \cref{fig:pipeline} for extracting six distinct visualizations for color and polarimetric data. Nonetheless, the raw data available in the PoTATO dataset enables the extraction of metrics through various approaches. In the subsequent sections, we elaborate on the main steps we implemented and the corresponding extracted visualizations.

\textit{STEP 1}: The RAW image is split into four $RGGB$ images, each corresponding to a specific polarization angle ($RGGB_{0}$, $RGGB_{45}$, $RGGB_{90}$, $RGGB_{135}$). This operation reduces the resolution by half to $1224 \times 1024$.

\textit{STEP 2}: The four $RGGB$ channels are debayered to generate the corresponding $RGB$ images ($RGB_{0}$, $RGB_{45}$, $RGB_{90}$, $RGB_{135}$).

\textit{STEP 3}: A color conversion is performed to extract grayscale images for each of the four polarization angles, providing channels ($M_{0}$, $M_{45}$, $M_{90}$, $M_{135}$). 

\textit{STEP 4}: Extracting the Stokes parameters is accomplished by applying \cref{eq:stokes} to each of the four channels (red, blue, green, monochrome), yielding four sets of Stokes vectors $(S_{R}, S_{G}, S_{B}, S_{M})$.

\textit{STEP 5}: The intensity parameters from the Stokes vectors ($S_{R_0}, S_{G_0}, S_{B_0}$) are used to generate the \textbf{RGB} visualization, providing a color representation of visible light that can be captured by regular cameras.

\textit{STEP 6}: By applying \cref{eq:dif} to the three Stokes vectors ($S_{R}, S_{G}, S_{B}$), the \textbf{DIF} representation is generated. This enables to filter out of specular reflections and focus on diffuse light. It is useful to reduce the bright reflections on the water's surface regardless of its angle of polarization, achieving greater flexibility compared to linear polarization filters fixed in front of the camera lens.

\textit{STEP 7}: The first Stokes parameter ($S_{M_0}$) is selected from the monochrome Stokes ($S_{M}$) to generate a visualization referred to as \textbf{MONO}. This representation shows the intensity of visible light in a grayscale image.

\textit{STEP 8}: The Angle of Linear Polarization (AoLP) is extracted from the monochrome Stokes vector $S_{M}$ using \cref{eq:aop}. This step assumes that the angle of polarization does not vary significantly based on color.

\textit{STEP 9}: The Degree of Linear Polarization (DoLP) is obtained by utilizing \cref{eq:dolp} on the monochrome Stokes vector ($S_{M}$). This step also assumes negligible variation of DoLP across different wavelengths. The DoLP values are then normalized between $0$ and $255$ and the color encoding presented in \cref{fig:color-encoding} is applied to generate the \textbf{DOLP} visualization.

\textit{STEP 10}: The DoLP and AoLP information are combined to generate a visualization referred to as \textbf{POL}. This image utilizes HSV encoding, with DoLP as the value, AoLP as the hue as shown in \cref{fig:color-encoding}, and saturation equal to 1. Such representation aids in visualizing the polarimetric data captured by the sensor, enabling rapid identification of the Angle of Linear Polarization, while simultaneously suppressing the meaningless and noisy AoLP information from regions where the light is not polarized.

\textit{STEP 11}: The \textbf{PAULI} visualization combines the $S_{m_{1}}$, $I_{45}$, and $S_{m_{0}}$ channels as $RGB$ images, mixing light intensity and polarimetric information. This approach draws inspiration from the Pauli decomposition method applied to polarimetric information from polarization-encoded SAR images as presented by Blin et al., \cite{blin2020new}.

\section{Experiments and Analysis}
\label{sec:experiments}

This section presents the experimental analysis using different object detection models. We show that off-the-shelf pre-trained object detection models don't perform well on this dataset and require fine-tuning. Experimental results indicate superiority of polarimetric-based inputs for medium and large objects, while RGB performs better for small objects. Finally, we also explore these observations through qualitative analysis and discuss the main challenges.

\subsection{Object Detection Baseline}

To establish a baseline for the PoTATO dataset, we conducted two sets of experiments: one using an off-the-shelf pre-trained model, and another using models fine-tuned specifically on our dataset. The evaluation employed commonly used metrics from the COCO object detection challenge \cite{lin2015microsoft} and all metrics were derived from the test set with 2000 images.

First, we experimented with YOLOv5 \cite{yolov5} using pre-trained weights on the COCO dataset \cite{lin2015microsoft} which already includes the \textit{bottle} class. The results in \cref{tab:od-coco-metrics-oftheshelf} indicate that the model trained on general object detection datasets does not generalize well for detecting floating bottles in the water surface, particularly when the objects appear small in the image. This underscores the unique challenges presented in our dataset and the need for more effective detection approaches. Notably, even without fine-tuning, the DIF channel shows significantly better performance when compared to other channels, suggesting that reducing reflections significantly enhances detection performance, particularly for large objects.

\begin{table}[thb]
    \caption{COCO Metrics (AP at IoU=.50:.05:.95) with pre-trained weights without fine-tuning. While the DIF channel yields the highest AP, particularly for large objects, overall detection performance remains limited.}
    \centering

    \begin{tabular}{l|r|r|rrr|rrrr}
    \toprule
    &\bf{Channel} &\bf{AP} &\bf{APs} &\bf{APm} &\bf{APl} &\bf{ARs} &\bf{ARm} &\bf{ARl} \\
    
    \midrule
    \multirow{6}{*}{\rotatebox[origin=c]{90}{\bf{Yolo-v5 m}}} 
    &MONO    &0.009 &0.000 &0.008 &0.150 &0.000 &0.001 &0.155 \\
    &RGB     &0.019 &0.000 &0.025 &0.350 &0.000 &0.022 &0.383 \\
    &DIF     &\textbf{0.025} &0.000 &\textbf{0.055} &\textbf{0.412} &0.000 &\textbf{0.052} &\textbf{0.440} \\
    &DOLP    &0.009 &0.000 &0.000 &0.025 &0.000 &0.000 &0.021 \\
    &POL     &0.005 &0.000 &0.004 &0.017 &0.000 &0.001 &0.017 \\
    &PAULI   &0.010 &0.000 &0.008 &0.126 &0.000 &0.007 &0.125 \\

    \bottomrule
    \end{tabular}

    \label{tab:od-coco-metrics-oftheshelf}
\end{table}

For the second set of experiments, we fine-tuned three models to each one of the six different channels of our dataset using the training set with 2000 images. They were evaluated them independently, and the results are presented in \cref{tab:od-coco-metrics-finetune}.

\begin{table}[thb]
    \caption{COCO Metrics (AP at IoU=.50:.05:.95) after fine tuning on PoTATO dataset. DIF and POL images have better results for medium and large bounding boxes. RGB image, performs better on small bounding boxes.}
    \centering

    \begin{tabular}{l|r|r|rrr|rrrr}
    \toprule
    &\bf{Channel} &\bf{AP} &\bf{APs} &\bf{APm} &\bf{APl} &\bf{ARs} &\bf{ARm} &\bf{ARl} \\
    
    \midrule
    \multirow{6}{*}{\rotatebox[origin=c]{90}{\bf{Faster R-CNN}}} 
    &MONO &0.448 &0.416 &0.555 &0.465 &0.476 &0.617 &0.493 \\
    &RGB &\bf{0.482} &\bf{0.446} &0.599 &0.483 &\bf{0.504} &0.651 &0.531 \\
    &DIF &0.477 &0.434 &0.609 &\bf{0.550} &0.492 &0.658 &\bf{0.609} \\
    &DOLP &0.418 &0.357 &0.592 &0.486 &0.424 &0.646 &0.539 \\
    &POL &0.428 &0.361 &\bf{0.627} &0.506 &0.426 &\bf{0.678} &0.579 \\
    &PAULI &0.446 &0.411 &0.557 &0.497 &0.468 &0.622 &0.529 \\
    
    \midrule
    \multirow{6}{*}{\rotatebox[origin=c]{90}{\bf{RetinaNet}}} 
    &MONO &0.362 &0.309 &0.512 &0.481 &0.424 &0.593 &0.520 \\
    &RGB &\bf{0.394} &\bf{0.323} &0.564 &0.523 &\bf{0.449} &0.633 &0.562 \\
    &DIF &0.388 &0.303 &0.593 &\bf{0.537} &0.423 &0.653 &0.575 \\
    &DOLP &0.321 &0.229 &0.560 &0.503 &0.352 &0.631 &0.558 \\
    &POL &0.345 &0.244 &\bf{0.597} &0.506 &0.367 &\bf{0.655} &\bf{0.584} \\
    &PAULI &0.368 &0.304 &0.537 &0.468 &0.419 &0.610 &0.496 \\

    \midrule
    \multirow{6}{*}{\rotatebox[origin=c]{90}{\bf{Yolo-v5 m}}} 
    &MONO &     0.402 &0.381 &0.500 &0.376 &0.466 &0.592 &0.455 \\
    &RGB &      \textbf{0.461} &\textbf{0.431} &0.578 &0.457 &\textbf{0.517} &0.662 &0.528 \\
    &DIF &   0.459 &0.425 &0.589 &\textbf{0.574} &0.509 &0.662 &\textbf{0.632} \\
    &DOLP &     0.400 &0.357 &0.560 &0.464 &0.442 &0.633 &0.566 \\
    &POL &      0.429 &0.386 &\textbf{0.597} &0.391 &0.478 &\textbf{0.674} &0.527 \\
    &PAULI &    0.450 &0.418 &0.572 &0.379 &0.504 &0.669 &0.453 \\
    \bottomrule
    \end{tabular}
    \label{tab:od-coco-metrics-finetune}
\end{table}

In addition to YOLOv5, we also included two other widely-used object detection models: Faster R-CNN and RetinaNet from Detectron2 \cite{wu2019detectron2}. These models were chosen for their diverse architectures: Faster R-CNN is two-stage detector with an efficient region proposal network, YOLOv5 represents a one-stage detection paradigm, and RetinaNet, another one-stage detector with a modified focal loss function. This variety enabled us to assess the consistency of performance differences across different input images. We used ResNet-50 \cite{he2016deep} with pre-trained weights as a backbone, and fine-tuned on our dataset. For consistency, we fine-tuned the three models with similar  parameters, including training for 100 epochs and batch size of 8. It is important to emphasize that augmentation was not used for preserving the correctness of the physics properties of light in the polarized channels. For single-channel images (MONO and DOLP) we used channel replication to match model input.

After fine-tuning on our dataset, substantial improvements were observed across all channels and models. The results highlight the advantages of polarimetric-based inputs, particularly POL and DIF images, which perform better in scenarios with medium and large bounding boxes. Conversely, RGB images excel when detecting small bounding boxes. Notably, the POL input, derived purely from polarimetric data, often surpasses RGB input, emphasizing the effectiveness of polarimetric modalities in object detection. It is worth noting that the AP metric, computed by averaging ten different IoU thresholds, is heavily influenced by the imbalanced distribution of bounding boxes sizes shown in \cref{fig:small-bboxes}.

Overall, our results indicate that the DIF channel, which simultaneously leverages chromatic and polarimetric information, is the most effective choice for detection applications. It demonstrates high performance across all bounding box sizes, making it particularly suitable for detecting bottles on the water surface.

\subsection{Advantages of Polarimetric Information}

\label{subsec:advantages}
Our experiments indicate that POL and DIF images show consistently increased performance for the images where the bottles are closer to the camera, while the RGB images provide better results when the bottles are farther away. Moreover, the accuracy gap between polarimetric-based and chromatic-based inputs increases with higher IoU thresholds, revealing that polarimetric data is more robust in precisely detecting the position of the bottles.

\begin{figure*}[hbt]
    \centering
    \includegraphics[width=\linewidth]{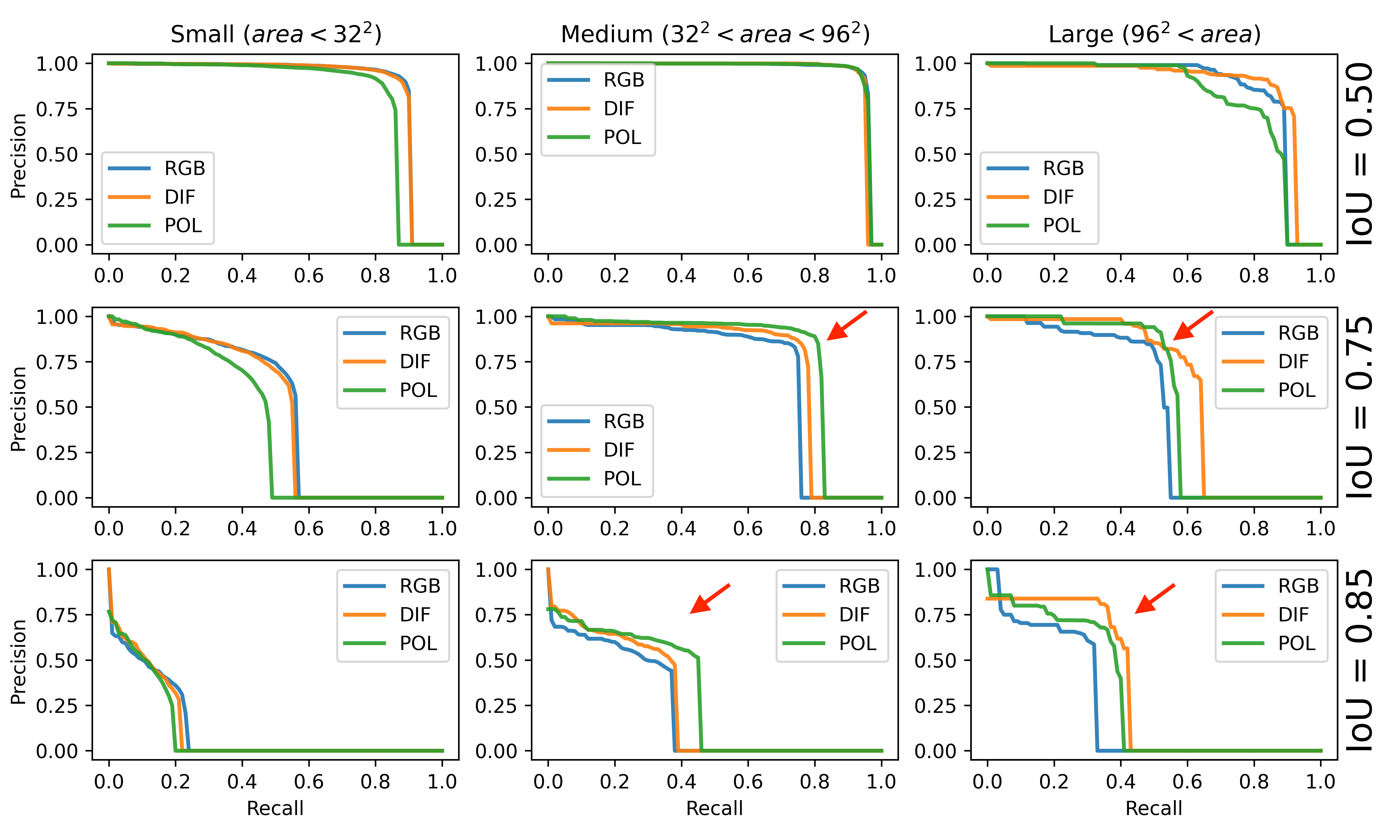}
    \caption{Precision-Recall curves for Faster R-CNN shows that polarimetric-based information (POL and DIF) presents better results in medium and large objects as the IoU threshold increases.}
    \label{fig:prcurves}
\end{figure*}

Such observation derives from \cref{fig:prcurves}, where we selected the three better performing channels: RGB, POL and DIF for the Faster R-CNN model and plotted the Precision-Recall curves for three different IoU thresholds. Faster R-CNN was selected due to its higher scores, and similar results are also observed on the other two models. The outcome is expected given that the region closer to the vessel has stronger polarimetric signals due to the height of the camera position and the Brewster's angle for the air/water interface explained in \cref{sec:theory}. In longer distances, the angle of incidence gets larger and the intensity of the polarimetric information decreases. In this scenario, the RGB tends to perform better for the detection of the bottles.

The relationship between bounding box size and distance is attributed to the position of the forward-facing camera fixed on the vessel. As a result, the bounding box size variation primarily derives from perspective geometry, with closer objects appearing larger on the image plane. 

\subsection{Qualitative Analysis}

\begin{figure}[bth]
\centering
\begin{subfigure}{0.32\linewidth}
    \includegraphics[width=\linewidth]{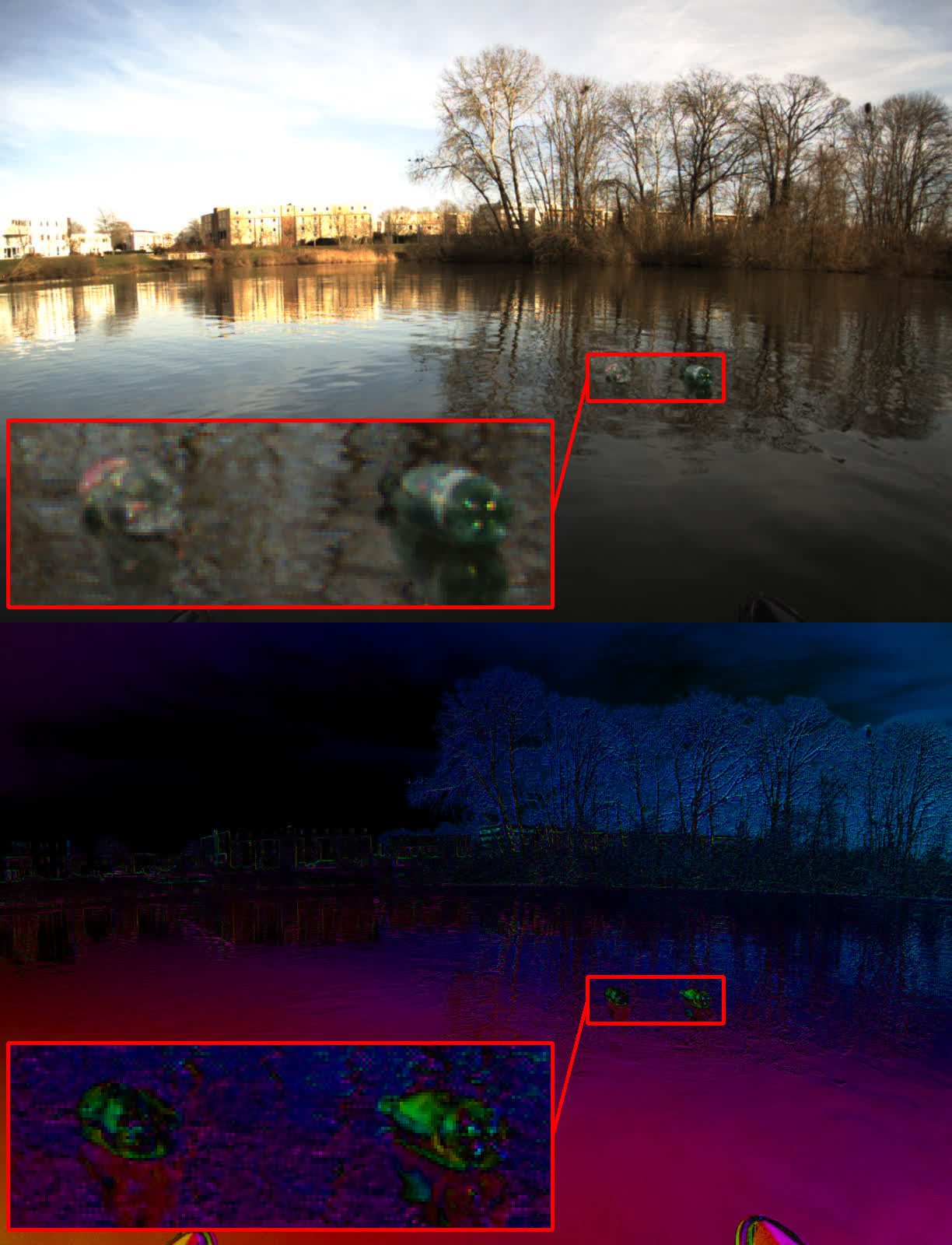}
    \caption{Object Concealed in Background Reflection.}
    \label{fig:q2}
\end{subfigure}
\hfill
\begin{subfigure}{0.32\linewidth}
    \includegraphics[width=\linewidth]{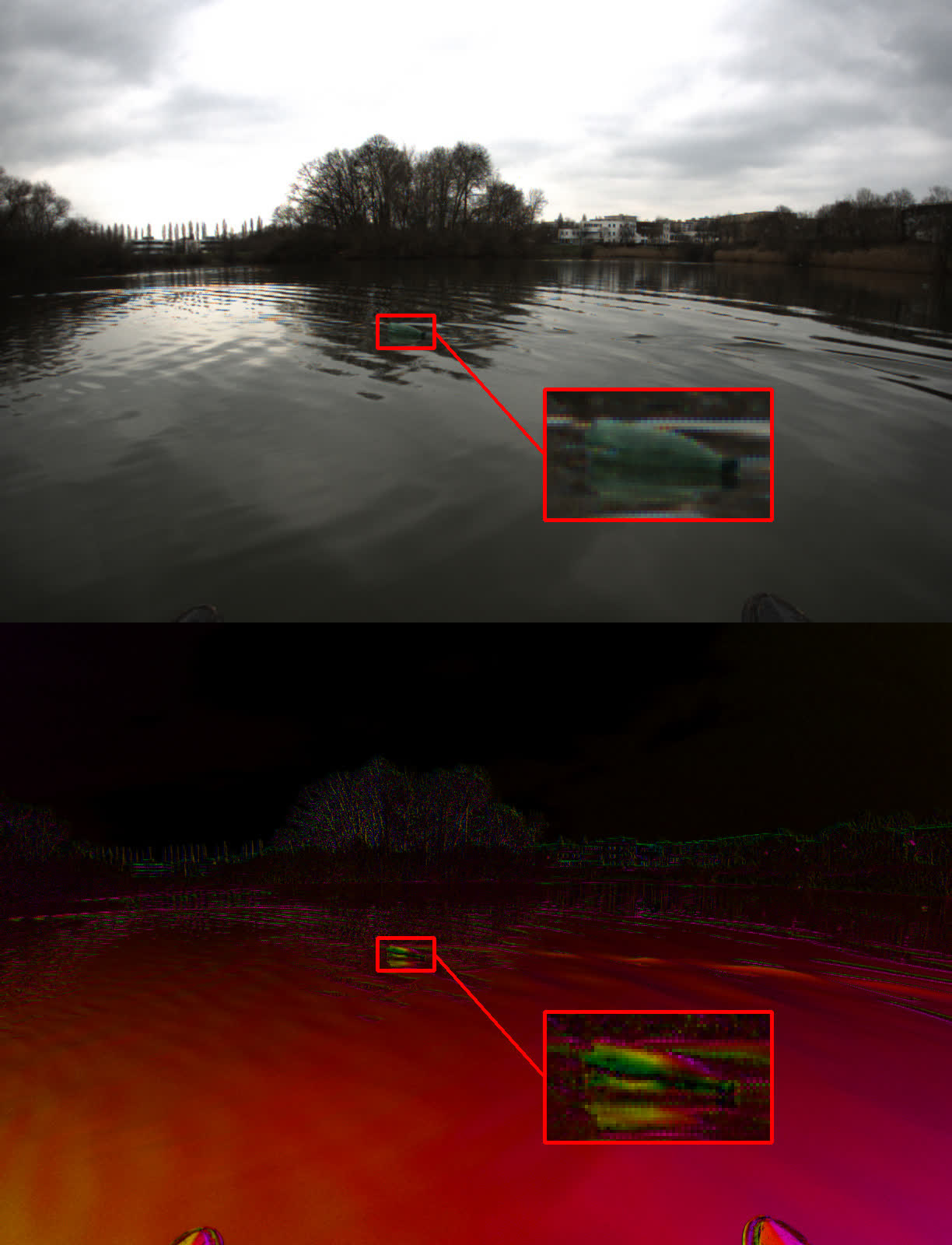}
    \caption{Color-based Concealing During a Cloudy Day.}
    \label{fig:q3}
\end{subfigure}
\hfill
\begin{subfigure}{0.32\linewidth}
    \includegraphics[width=\linewidth]{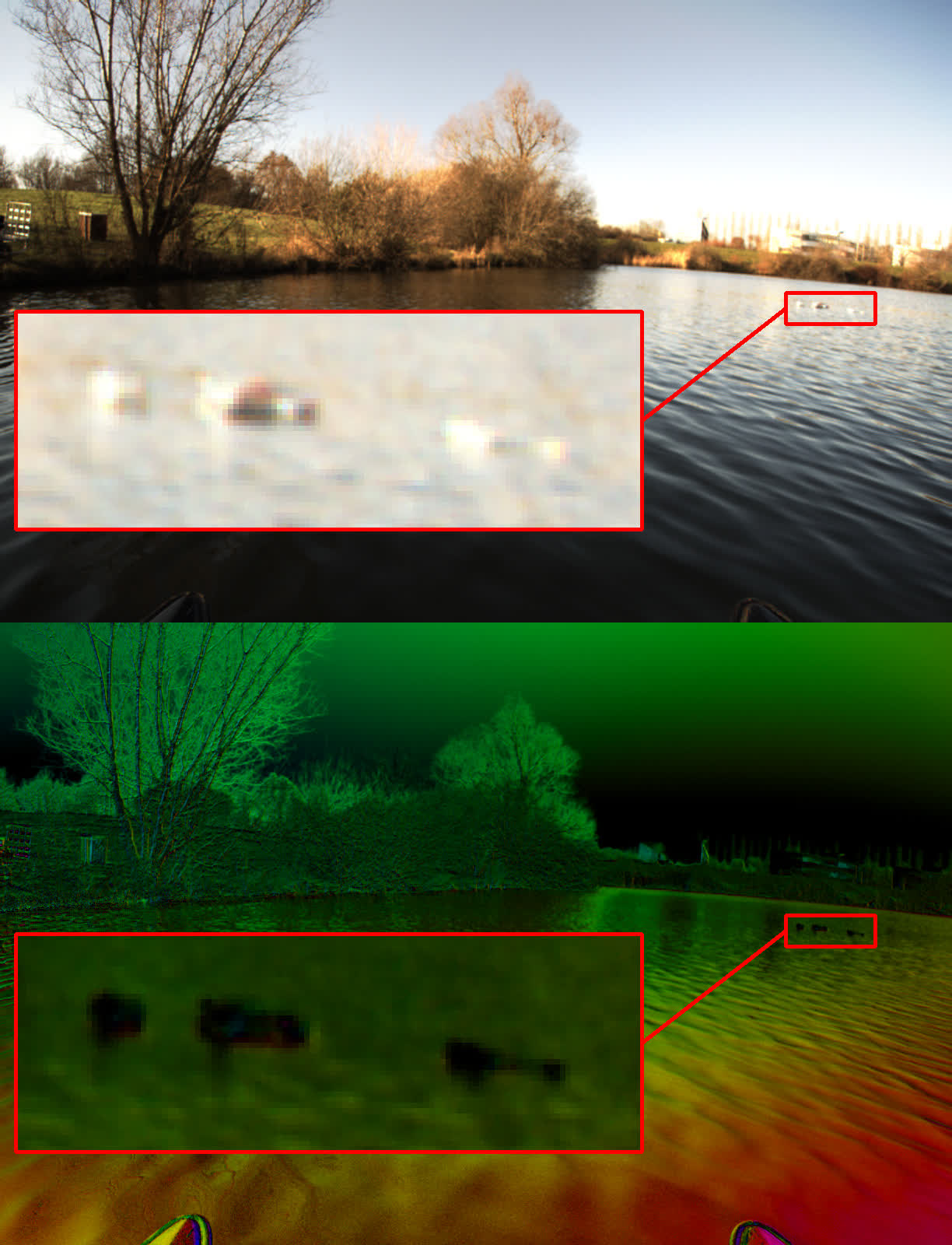}
    \caption{High Brightness and skylight reflection on a sunny day.}
    \label{fig:q1}
\end{subfigure}
\caption{Qualitative Analysis}
\label{fig:figures}
\end{figure}

To illustrate the difference in the features conveyed by the polarimetric and chromatic images, three images were selected. First, \Cref{fig:q2} showcases bottles floating amidst overlapping tree reflections. In the RGB image, it is challenging to distinguish it from the background. However, in the corresponding POL image, the distinct Angle of Linear Polarization (AoLP) is emphasized, exhibiting stronger contrast in a different color. Second, \Cref{fig:q3} presents a green bottle nearly concealed by the water with a similar color while the POL image effectively accentuates the contrasting AoLP. This image was captured on a cloudy day, where the skylight polarization is blocked by the clouds and the predominant source of polarization is the reflection on the water's surface. Third, \Cref{fig:q1} illustrates a scenario where plastic bottles are situated in a well-lit area of the lake. The RGB image captures intense light reflection, indicating the presence of an object, but fails to clearly delineate its boundaries. The corresponding POL image shows sharply delineated edges of the bottles, highlighting the contrast with the water surface. 
% It is also possible to observe that this image was acquired on a sunny day, due to the high polarization in the sky that is also reflected on the water.

\subsection{Challenges}

\begin{figure}[tb]
\begin{center}
\includegraphics[width=0.8\linewidth]{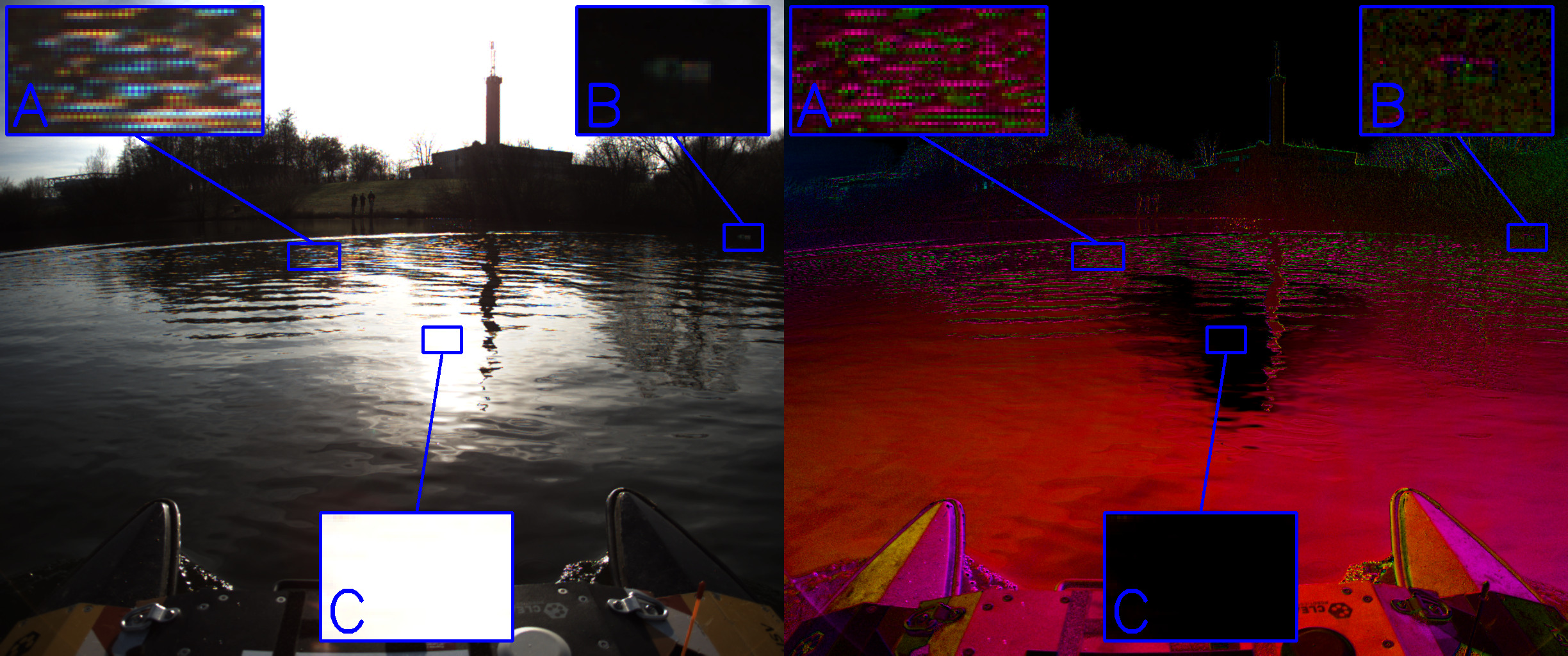}
\end{center}
   \caption{Main Challenges: A: Artifacts; B: Small Object; C: Saturation}
\label{fig:challenges}
\end{figure}

Three main challenges were identified in the PoTATO dataset and illustrated in \cref{fig:challenges}. First, is the large amount of images where the far-away bottles appear small in the image. The dataset contains a considerable amount of bounding boxes that have dimensions lower than ${14}^2$ pixels, posing a challenge for object detection. Second, is a correlated problem deriving from the fact that the superpixel shown in \cref{fig:sensor} is quite large on the image plane, creating artifacts, especially on edges and image areas with high frequency which is particularly common due to patterns created by ripples in the water surface. Both these challenges can be mitigated with different strategies for extracting the channels from the raw data \cite{qiu2021linear} and can become the scope of further research. The third challenge was the saturation of the sensor under strong light reflection. In such cases, all polarization channels were measuring maximum intensity without any relative difference between them, making it impossible to estimate the degree and angle of polarization. Beyond using auto-brightness function, this could be mitigated by using High Dynamic Range (HDR) to capture two consecutive images with shorter and longer exposure to capture data on bright and dark regions. This approach was avoided to prevent temporal misalignment between image pairs.

\section{Conclusion}

In our study, we demonstrate the potential of polarimetric images to enhance vision system performance, particularly within the scope of autonomous water-cleaning robots operating in challenging outdoor environments. Our quantitative analysis shows that polarimetric images are capable of outperforming chromatic images for the object detection task in regions with heightened polarization properties. Through our initial experiments, we offer theoretical and practical insights into the advantages and challenges of using microgrid-polarization sensors for object detection in environments where light reflections on water surfaces are abundant. By providing the first dataset with raw polarimetric images and accompanying code we aim to inspire further research and enable the development of novel approaches for fusing color and polarization modalities, with a strong potential for advancing the state-of-the-art perception algorithms.

\textbf{Acknowledgments}: This research was supported by the French Agence Nationale de la Recherche under grant ANR-23-CE23-0030 (Project R3AMA).

\bibliographystyle{splncs04}
\bibliography{main}

\begin{thebibliography}{10}
\providecommand{\url}[1]{\texttt{#1}}
\providecommand{\urlprefix}{URL }
\providecommand{\doi}[1]{https://doi.org/#1}

\bibitem{andreou2002polarization}
Andreou, A.G., Kalayjian, Z.K.: Polarization imaging: principles and integrated
  polarimeters. IEEE Sensors journal  \textbf{2}(6),  566--576 (2002)

\bibitem{blin2019adapted}
Blin, R., Ainouz, S., Canu, S., Meriaudeau, F.: Adapted learning for
  polarization-based car detection. In: Fourteenth International Conference on
  Quality Control by Artificial Vision. vol. 11172, pp. 312--318. SPIE (2019)

\bibitem{blin2020new}
Blin, R., Ainouz, S., Canu, S., Meriaudeau, F.: A new multimodal rgb and
  polarimetric image dataset for road scenes analysis. In: Proceedings of the
  IEEE/CVF Conference on Computer Vision and Pattern Recognition Workshops. pp.
  216--217 (2020)

\bibitem{blin2021multimodal}
Blin, R., Ainouz, S., Canu, S., Meriaudeau, F.: Multimodal polarimetric and
  color fusion for road scene analysis in adverse weather conditions. In: 2021
  IEEE International Conference on Image Processing (ICIP). pp. 3338--3342.
  IEEE (2021)

\bibitem{blin2021polarlitis}
Blin, R., Ainouz, S., Canu, S., Meriaudeau, F.: The polarlitis dataset: Road
  scenes under fog. IEEE Transactions on Intelligent Transportation Systems
  \textbf{23}(8),  10753--10762 (2021)

\bibitem{chamas2020degradation}
Chamas, A., Moon, H., Zheng, J., Qiu, Y., Tabassum, T., Jang, J.H., Abu-Omar,
  M., Scott, S.L., Suh, S.: Degradation rates of plastics in the environment.
  ACS Sustainable Chemistry \& Engineering  \textbf{8}(9),  3494--3511 (2020)

\bibitem{cheng2021flow}
Cheng, Y., Zhu, J., Jiang, M., Fu, J., Pang, C., Wang, P., Sankaran, K.,
  Onabola, O., Liu, Y., Liu, D., et~al.: Flow: A dataset and benchmark for
  floating waste detection in inland waters. In: Proceedings of the IEEE/CVF
  International Conference on Computer Vision. pp. 10953--10962 (2021)

\bibitem{dave2022pandora}
Dave, A., Zhao, Y., Veeraraghavan, A.: Pandora: Polarization-aided neural
  decomposition of radiance. In: Computer Vision--ECCV 2022: 17th European
  Conference, Tel Aviv, Israel, October 23--27, 2022, Proceedings, Part VII.
  pp. 538--556. Springer (2022)

\bibitem{de2021quantifying}
De~Vries, R., Egger, M., Mani, T., Lebreton, L.: Quantifying floating plastic
  debris at sea using vessel-based optical data and artificial intelligence.
  Remote Sensing  \textbf{13}(17), ~3401 (2021)

\bibitem{van2020plastic}
van Emmerik, T., Schwarz, A.: Plastic debris in rivers. Wiley Interdisciplinary
  Reviews: Water  \textbf{7}(1),  e1398 (2020)

\bibitem{foster2018polarisation}
Foster, J.J., Temple, S.E., How, M.J., Daly, I.M., Sharkey, C.R., Wilby, D.,
  Roberts, N.W.: Polarisation vision: overcoming challenges of working with a
  property of light we barely see. The Science of Nature  \textbf{105},  1--26
  (2018)

\bibitem{gao2022polarimetric}
Gao, D., Li, Y., Ruhkamp, P., Skobleva, I., Wysocki, M., Jung, H., Wang, P.,
  Guridi, A., Busam, B.: Polarimetric pose prediction. In: Computer
  Vision--ECCV 2022: 17th European Conference, Tel Aviv, Israel, October
  23--27, 2022, Proceedings, Part IX. pp. 735--752. Springer (2022)

\bibitem{goldstein2017polarized}
Goldstein, D.H.: Polarized light. CRC press (2017)

\bibitem{he2016deep}
He, K., Zhang, X., Ren, S., Sun, J.: Deep residual learning for image
  recognition. In: Proceedings of the IEEE conference on computer vision and
  pattern recognition. pp. 770--778 (2016)

\bibitem{iqbal2022object}
Iqbal, A., Garcia, M.G., Chellappan, L., Gans, N.: Object detection and
  classification for small objects in/on water. Journal of Electronic Imaging
  \textbf{31}(3),  033041--033041 (2022)

\bibitem{jia2023deep}
Jia, T., Kapelan, Z., de~Vries, R., Vriend, P., Peereboom, E.C., Okkerman, I.,
  Taormina, R.: Deep learning for detecting macroplastic litter in water
  bodies: a review. Water Research p. 119632 (2023)

\bibitem{kalra2020deep}
Kalra, A., Taamazyan, V., Rao, S.K., Venkataraman, K., Raskar, R., Kadambi, A.:
  Deep polarization cues for transparent object segmentation. In: Proceedings
  of the IEEE/CVF Conference on Computer Vision and Pattern Recognition. pp.
  8602--8611 (2020)

\bibitem{lei2020polarized}
Lei, C., Huang, X., Zhang, M., Yan, Q., Sun, W., Chen, Q.: Polarized reflection
  removal with perfect alignment in the wild. In: Proceedings of the IEEE/CVF
  Conference on Computer Vision and Pattern Recognition. pp. 1750--1758 (2020)

\bibitem{lin2015microsoft}
Lin, T.Y., Maire, M., Belongie, S., Bourdev, L., Girshick, R., Hays, J.,
  Perona, P., Ramanan, D., Zitnick, C.L., Dollár, P.: Microsoft coco: Common
  objects in context (2015)

\bibitem{macleod2021global}
MacLeod, M., Arp, H.P.H., Tekman, M.B., Jahnke, A.: The global threat from
  plastic pollution. Science  \textbf{373}(6550),  61--65 (2021)

\bibitem{politikos2023using}
Politikos, D.V., Adamopoulou, A., Petasis, G., Galgani, F.: Using artificial
  intelligence to support marine macrolitter research: A content analysis and
  an online database. Ocean \& Coastal Management  \textbf{233},  106466 (2023)

\bibitem{proencca2020taco}
Proen{\c{c}}a, P.F., Simoes, P.: Taco: Trash annotations in context for litter
  detection. arXiv preprint arXiv:2003.06975  (2020)

\bibitem{qiu2021linear}
Qiu, S., Fu, Q., Wang, C., Heidrich, W.: Linear polarization demosaicking for
  monochrome and colour polarization focal plane arrays. In: Computer Graphics
  Forum. vol.~40, pp. 77--89. Wiley Online Library (2021)

\bibitem{ratliff2009interpolation}
Ratliff, B.M., LaCasse, C.F., Tyo, J.S.: Interpolation strategies for reducing
  ifov artifacts in microgrid polarimeter imagery. Optics express
  \textbf{17}(11),  9112--9125 (2009)

\bibitem{ratliff2011detection}
Ratliff, B.M., LeMaster, D.A., Mack, R.T., Villeneuve, P.V., Weinheimer, J.J.,
  Middendorf, J.R.: Detection and tracking of rc model aircraft in lwir
  microgrid polarimeter data. In: Polarization Science and Remote Sensing V.
  vol.~8160, pp. 29--41. SPIE (2011)

\bibitem{Label_Studio}
Tkachenko, M., Malyuk, M., Holmanyuk, A., Liubimov, N.: {Label Studio}: Data
  labeling software (2020-2022),
  \url{https://github.com/heartexlabs/label-studio}, open source software
  available from https://github.com/heartexlabs/label-studio

\bibitem{yolov5}
Ultralytics: {ultralytics/yolov5: v7.0 - YOLOv5 SOTA Realtime Instance
  Segmentation}. \url{https://github.com/ultralytics/yolov5.com} (2022).
  \doi{10.5281/zenodo.7347926}, \url{https://doi.org/10.5281/zenodo.7347926},
  accessed: 7th May, 2023

\bibitem{wang2018bottle}
Wang, J., Guo, W., Pan, T., Yu, H., Duan, L., Yang, W.: Bottle detection in the
  wild using low-altitude unmanned aerial vehicles. In: 2018 21st International
  Conference on Information Fusion (FUSION). pp. 439--444. IEEE (2018)

\bibitem{wang2022principle}
Wang, Y., Su, Y., Sun, X., Hao, X., Liu, Y., Zhao, X., Li, H., Zhang, X., Xu,
  J., Tian, J., et~al.: Principle and implementation of stokes vector
  polarization imaging technology. Applied Sciences  \textbf{12}(13), ~6613
  (2022)

\bibitem{wu2019detectron2}
Wu, Y., Kirillov, A., Massa, F., Lo, W.Y., Girshick, R.: Detectron2.
  \url{https://github.com/facebookresearch/detectron2} (2019)

\bibitem{zhang2021survey}
Zhang, R., Li, S., Ji, G., Zhao, X., Li, J., Pan, M.: Survey on deep
  learning-based marine object detection. Journal of Advanced Transportation
  \textbf{2021},  1--18 (2021)

\bibitem{zou2023object}
Zou, Z., Chen, K., Shi, Z., Guo, Y., Ye, J.: Object detection in 20 years: A
  survey. Proceedings of the IEEE  (2023)

\end{thebibliography}

%\maketitle

\end{document}